\documentclass{article}

\usepackage{microtype}
\usepackage{graphicx}
\usepackage{subcaption}
\usepackage{booktabs} 

\usepackage{amsmath,amsfonts,bm}

\def\eqref#1{equation~\ref{#1}}

\def\1{\bm{1}}

\def\vz{{\bm{z}}}

\def\mI{{\bm{I}}}

\def\mT{{\bm{T}}}

\DeclareMathAlphabet{\mathsfit}{\encodingdefault}{\sfdefault}{m}{sl}
\SetMathAlphabet{\mathsfit}{bold}{\encodingdefault}{\sfdefault}{bx}{n}

\usepackage{hyperref}
\usepackage{url}
\usepackage{epsfig,parskip,setspace,tabularx,xspace,amsmath}

\usepackage[accepted]{icml2019}

\icmltitlerunning{Unsupervised and interpretable scene discovery with Discrete-Attend-Infer-Repeat}

\usepackage[textsize=scriptsize]{todonotes}
\definecolor{mysidenotefgcolor}{rgb}{0,0,0}
\definecolor{mysidenotefgauthorcolor}{rgb}{0.5,0.5,0.5}

\begin{document}

\twocolumn[
\icmltitle{Unsupervised and interpretable scene discovery \\ with Discrete-Attend-Infer-Repeat}



\icmlsetsymbol{equal}{*}

\begin{icmlauthorlist}
\icmlauthor{Duo Wang}{cam}
\icmlauthor{Mateja Jamnik}{cam}
\icmlauthor{Pietro Lio}{cam}

\end{icmlauthorlist}

\icmlaffiliation{cam}{Department of Computer Science and Technology, University of Cambridge, Cambridge, UK}

\icmlcorrespondingauthor{Duo Wang}{duo.wang@cl.cam.ac.uk}

\icmlkeywords{Machine Learning, ICML, Unsupervised Learning}

\vskip 0.3in
]



\printAffiliationsAndNotice{\icmlEqualContribution} 

\begin{abstract}
In this work we present Discrete Attend Infer Repeat (Discrete-AIR), a Recurrent Auto-Encoder with structured latent distributions containing discrete categorical distributions, continuous attribute distributions, and factorised spatial attention. While inspired by the original AIR model~\cite{eslami2016attend} andretaining AIR model's capability in identifying objects in an image, Discrete-AIR provides direct interpretability of the latent codes. We show that for Multi-MNIST~\cite{eslami2016attend} and a multiple-objects version of dSprites~\cite{dsprites17} dataset, the Discrete-AIR model needs just one categorical latent variable, one attribute variable (for Multi-MNIST only), together with  spatial attention variables, for efficient inference. We perform analysis to show that the learnt categorical distributions effectively capture the categories of objects in the scene for Multi-MNIST and for Multi-Sprites. 
\end{abstract}

\section{Introduction}

Many real-world tasks, such as inferring physical relationships between objects in an image and visual-spatial reasoning, requires identifying and learning a useful representation of elements in the scene. Such objects can be conveniently encoded into a representation containing its category, attributes and spatial positions and orientations. For example, an object can be of category vehicle, with attributes such as red colour and 4 doors, and positioned at the bottom right of the scene in a specific orientation. Humans, when recognising objects or trying to draw them, are believed to have attentional templates~\cite{carlisle2011attentional} of different categories of objects in mind that are augmented by different attributes and selected spatially via attention. 

\begin{figure}[h]
	\begin{center}
		\includegraphics[width=0.9\linewidth]{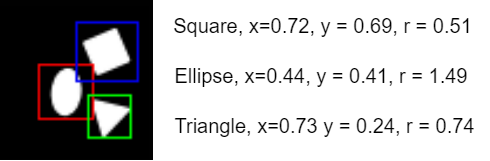}
	\end{center}
	\caption{Illustration of encoding scenes into category and attribute latent code.  From a scene containing three different shapes, Discrete-AIR separately identifies each of the shapes (with different coloured bounding box). It also estimates spatial x-axis and y-axis locations, and orientation of the shape. }
	\label{fig:discrete_illu}
\end{figure}

Machine approaches to such problems often use generative models such as Variational Auto-Encoder (VAE)~\cite{kingma2013auto}, which use an inference model to infer latent codes corresponding to the representation, and a generative model which reconstruct data given the representation. Recurrent versions of VAE such as Attend-Infer-Repeat (AIR) model by Eslami et al~\cite{eslami2016attend} have been developed to decompose a scene into multiple objects with each represented by latent code $z = (z_{what},z_{where},z_{pres})$. While this latent code disentangles spatial information $z_{where}$ and object presence $z_{pres}$,  for most of the tasks, the object representation $z_{what}$ is an entangled real-valued vector and thus difficult to interpret. While AIR does propose the possibility of using discrete latent code as $z_{what}$, it only experimented with the discrete code with specifically designed graphics-engine as decoder. Here, we propose Discrete-AIR, an end-to-end trainable autoencoder which structures latent representation $z_{what}$ into $z_{cat}$ representing category of objects and $z_{attr}$ representing attributes of objects. Figure~\ref{fig:discrete_illu} illustrates how a scene of different shapes can be separately identified into different categories with varying attributes. This decomposition is similar to InfoGAN by Chen et al~\cite{chen2016infogan}, which also decompose representation into style and shape using a modified Generative Adversarial Network~\cite{goodfellow2014generative}. However, there are two main differences. Firstly, while InfoGAN uses a mutual information objective in addition to GAN objective to encourage disentangled coding, Discrete-AIR only uses the Variational Lower Bound (ELBO) objectives and encourages disentanglement through the inductive-bias structure of latent code. Secondly, while InfoGAN is only applied to images containing single objects, Discrete-AIR is developed for scenes with multiple objects. 

Related to this work are other approaches which decompose scenes into different categories. Neural Expectation Maximization (NEM) by Greff et al~\cite{greff2017neural} implemented Expectation-Maximization algorithm with an end-to-end trainable neural network. NEM is able to perceptually group pixels of an image into different clusters. However, it does not learn a generative model that allows controllable generation like using $z_{cat}$ and $z_{where}$ in Discrete-AIR. Ganin et al~\cite{ganin2018synthesizing} train a neural network to synthesise programs that can be fed into a graphics engine to generate scenes. While it learns an inference model for the generative model, a graphics engine that can provide learning gradients is pre-defined and not learnt. In contrast, Discrete-AIR jointly learns an inference model and a generative model from scratch.

We show that our Discrete-AIR model can decompose scenes into a set of interpretable latent codes for two multi-object datasets, namely Multi-MNIST dataset as used in the original AIR model~\cite{eslami2016attend} and a multi-object dataset in similar style as the dSprites dataset~\cite{dsprites17}. We show that unsupervised training of Discrete-AIR model is able to effectively capture the categories of objects in the scene for Multi-MNIST and for Multi-Sprites datasets.

\section{Attend Infer Repeat}
Attend-Infer-Repeat (AIR) model, introduced by Eslami et al~\cite{eslami2016attend}, is a recurrent version of Variational Auto-Encoder (VAE)~\cite{kingma2013auto} that decomposes a scene into multiple objects represented by latent code $z^i = (z^i_{what}, z^i_{where}, z^i_{pres})$ at each recurrent time step $i$. Among them $z^i_{pres}$ is a binary discrete variable encoding whether an object is inferred in current step $i$. If $z^i_{pres}$ is 0, the inference will be stopped. The sequence of $z^i_{pres}$ for all $i$ can be concatenated into a vector of $n$ ones and a final zero. $n$ therefore is a variable representing the number of objects in the scene. $z^i_{where}$ is a spatial attention parameter used to locate a target object in the image, and $z^i_{what}$ is the latent code of the target object. In AIR an amortised variational approximation $q_{\phi}(z|x)$, as computed in equation~\ref{eq:q}, is used to approximate true posterior $p(z|x)$ by minimising KL divergence $KL[q_{\phi}(z|x)||(z|x)]$. In AIR implementation, $z_{what}$ and $z_{where}$ are parametrised as Gaussian distributions with diagonal covariance $\mathcal{N}(\mu,\Sigma)$.

\begin{equation}\label{eq:q}
q_{\phi}(z|x) = q(z^{n+1}_{pres}|z^{1:n},x) \prod_{i=1}^{n} q_{\phi}(z^i,z^i_{pres} = 1 |x,z^{1:i-1})
\end{equation}

In the generative model of AIR, the number of objects $n$ can be sampled from a prior such as geometric prior, and then form the sequence of $z^i_{pres}$. Next, $z^i_{what}$ and $z^i_{where}$ are sampled from $N(0,\mI)$. An object $o^i$ is generated by processing $z^i_{what}$ through a decoder. $o^i$ is then written to the canvas, gated by $z^i_{pres}$ and with scaling and translation specified by $z^i_{where}$ using Spatial Transformer~\cite{jaderberg2015spatial}, a powerful spatial attention module. The generative model can be summarised in equation~\ref{eq:generative}, where $f_{dec}$ is the decoder, $ST$ is the spatial transformer and $\odot$ is element-wise product.

\begin{flalign}\label{eq:generative}
& p_{\theta}(x|z) = \mathcal{N}(x|y,\sigma_xI) \\ 
& y = \sum_{i=1}^{n} ST(f_{dec}(z^i_{what}),z^i_{where}) \odot z^i_{pres} 
\end{flalign}
Inference and generative models of AIR are jointly optimised by maximising the lower bound $\mathcal{L}(q_{\phi},p_{\theta}) = E_{q_{\phi}}[\log \frac{p_{\theta}(x,z,n)}{q_{\phi}(z,n|x)}]$. While sampling operation of $z$ is not differentiable (which is a requirement for gradient-based training), there are various ways to circumvent this. For the continuous latent codes, re-parametrisation trick for VAE~\cite{kingma2013auto} is applied, which lets parameters estimated from the inference model to deterministically modify a sampled distribution, thereby allowing back-propagation through the deterministic function. For discrete latent codes, AIR uses NVIL likelihood ratio estimator introduced by Mnih et al~\cite{mnih2014neural} to produce an unbiased estimate of the gradient for discrete latent variables.

\section{Discrete-AIR}

While the AIR model can encode objects in a scene into latent code $z_{what}$, the representation is still entangled and therefore not interpretable. In Discrete-AIR, we introduce structure into the latent distribution to encourage disentanglement. We break $z_{what}$ into $z_{cat}$ and $z_{attr}$. $z_{cat}$ is discrete latent variable that captures the category of the object, while $z_{attr}$ is a combination of  continuous and discrete latent variables that captures attributes of the object. We do not use any objective function to encourage $z_{cat}$ to capture category and $z_{attr}$ to capture attributes. Rather, we allow the model to automatically learn the best way of using these discrete and latent variables through the process of likelihood maximisation. 

\subsection{Sampling discrete variable}\label{sec:discrete}

In Discrete-AIR, we treat binary discrete variables as scalar of 0/1 values and multi-class categorical discrete variables as one-hot vectors. As sampling from a discrete distribution is non-differentiable, we model discrete latent variables with Gumbel Softmax~\cite{maddison2016concrete,jang2016categorical}, a continuous approximation to the discrete distribution from which we can sample approximately one-hot discrete vector $y$ where:

\begin{equation}\label{eq:gumbel_softmax}
y_i = \frac{exp(\frac{log \, a_i+g_i}{\tau})}{\sum_{j=1}^{k} exp(\frac{ log \, a_i+g_i}{\tau})} \quad ;i=1,\dots,k
\end{equation}

$a_i$ are a parametrisation of the distribution, $g_i$ are Gumbel noise sampled from the Gumbel distribution $Gumbel(0,1)$, and $\tau$ is temperature parameters controlling smoothness of the distribution. As $\tau \rightarrow 0$, the distribution converges to a discrete distribution. For binary discrete variables such as $z_{pres}$, we use Gumbel Sigmoid, which is essentially Gumbel softmax with softmax function replaced with Sigmoid function:

\begin{equation}\label{eq:gumbel_sigmoid}
y = \frac{exp(\frac{log \, a +g }{\tau})}{1 + exp(\frac{ log \, a+g}{\tau})} 
\end{equation}

In contrast to the NVIL estimator~\cite{mnih2014neural} used in the original AIR model, we found that Gumbel softmax/Sigmoid is more stable during training, experiencing no model collapse during all the training experiments.

\subsection{Generative model}\label{sec:gen_model}

\begin{figure}[h]
	\begin{center}
		\includegraphics[width=0.95\linewidth]{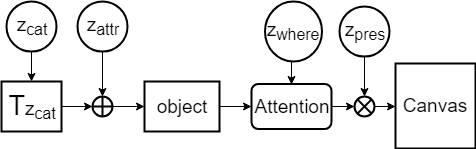}
	\end{center}
	\caption{Generative Model of Discrete-AIR.}
	\label{fig:generative}
\end{figure}

The probabilistic generative model is shown in figure~\ref{fig:generative}. From $z_{cat}$, a template $T_{z_{cat}}$ of this object category is generated. This template is then modified by attributes $z_{attr}$ into an image of object $o$ that is subsequently drawn onto the canvas using spatial write attention. $z_{cat}$, $z_{attr}$, $z_{where}$ and $z_{pres}$, jointly as $\vz_t$, are estimated from the inference model for each time step $t$ of inference. 

We replace the decoder function $f_{dec}(z^i_{what})$ from the original AIR model with a new function $f_{dec}(z^i_{cat},z^i_{attr})$ parametrised by category variable $z_{cat}$ and $z_{attr}$. There are various candidate functions for combining $z_{cat}$ and $z_{attr}$. We have experimented with three different variations, which are:
\begin{itemize}
\item \textbf{Additive}: $f_{dec}(z^i_{cat},z^i_{attr}) = f(f_t(z^i_{cat})+f_a(z^i_{attr}))$ where $f_t(z^i_{cat})$ generates a template, while $f_a(z^i_{attr})$ generates an additive modification of template.
\item \textbf{Multiplicative}: $f_{dec}(z^i_{cat},z^i_{attr}) = f(f_t(z^i_{cat}) \odot f_a(z^i_{attr}))$ where $f_t(z^i_{cat})$ generates a template, while $f_a(z^i_{attr})$ generates a multiplicative modification of template.
\item \textbf{Convolutional}: $f_{dec}(z^i_{cat},z^i_{attr}) = f(f_{t(conv)}(z^i_{cat}) * f_{a(filter)}(z^i_{attr}))$ where $f_{t(conv)}(z^i_{cat})$ generates a template, while $f_{a(filter)}(z^i_{attr})$ generates a set of convolution kernels that can be convolved with template to modify it.
\end{itemize}

In our experiments we found that the choice of combining functions has only  a small effect on the model performance. We found that the additive combining function performs slightly better, and thus use this function in all of the experiments presented.

In the original AIR model, the spatial transformation operation specified by attention variable $z_{where}$ only contains translation and scaling. Affine transformations such as rotation and shearing are accounted for in the latent variable in an entangled way. In Discrete-AIR, we explicitly introduce additional spatial transformer networks that account for rotation and skewing, thereby allowing $z_{attr}$ to have a reduced number of variables. The spatial attention for the generative decoder is thus factorised as in equation~\ref{eq:st},

\begin{flalign}\label{eq:st}
&\mT^d = \mT^d_{st} \mT^d_r \mT^d_k = \\\nonumber
&\begin{bmatrix}
s_x & 0 &t_x \\
0 & s_y & t_y\\
0 & 0 & 1 
\end{bmatrix}
\begin{bmatrix}
cos(\omega) & sin(-\omega) & 0 \\
sin(\omega) & cos(\omega) & 0\\
0 & 0 & 1 
\end{bmatrix}
\begin{bmatrix}
1+k_x k_y & k_x &0 \\
k_y & 1 & 0\\
0 & 0 & 1 
\end{bmatrix}
\end{flalign}

 where $\mT^d_{st}$ is the combined transformation matrix of translation and scaling used in the original AIR model, $\mT_r$ is the transformation matrix for rotation and $\mT_k$ is the transformation matrix for skewing. In the matrix, $s_x$ and $s_y$ are for scaling, $t_x$ and $t_y$ are horizontal and vertical translations, $\omega$ is an angle of rotation, $k_x$ and $k_y$ are parameters for shearing in horizontal and vertical axis.

\subsection{Inference}
\begin{figure*}[h]
	\begin{center}
		\includegraphics[width=0.95\linewidth]{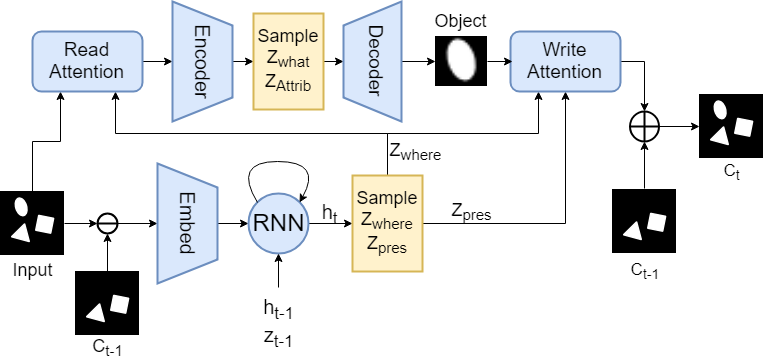}
	\end{center}
	\caption{Overview of Discrete-AIR architecture. Blue parts are neural-network trainable modules and yellow parts are sampling processes.}
	\label{fig:overview}
\end{figure*}

Figure~\ref{fig:overview} shows an overview of the Discrete-AIR architecture. At inference step $t$, a difference image between input image $D$ and previous canvas $C_{t-1}$ is fed together with previous latent code $\vz_{t-1}$ into a Recurrent Neural Network (RNN), implemented as Long Short-Term Memory (LSTM)~\cite{hochreiter1997long} to generate parameters of the distribution for $z_{where}$ and $z_{pres}$. Spatial Attention module then attends to parts of the image and applies transformation according to $z_{where}$. We enforce the encoding transformation $\mT^e$ to be the inverse of decoding transformation $\mT^d$, which means $\mT^e \mT^d = \mI$. This constraint forces the model to match attended objects in the scene with the invariant template specified by $z_{cat}$. In practice, we compute $\mT^e$ as the product of inverses of the transformation matrices composing $\mT^d$:
\begin{equation}
\mT^e = {\mT^{d}}^{-1} = {\mT^d_{k}}^{-1} {\mT^d_r}^{-1} {\mT^d_{st}}^{-1}
\end{equation}

The transformed image is then processed by an encoder to estimate parameters of distributions for $z_{cat}$ and $z_{attr}$. $z_{cat}$ are sampled from Gumbel Softmax as discussed in section~\ref{sec:discrete}. $z_{attr}$ can be sampled from any distribution that is suitable for the paradigm of tasks. For tasks presented, continuous variables such as the colour intensity or part deformation of an object can be sampled from a multi-variate Gaussian distribution using the Re-parameterization trick~\cite{kingma2013auto}, which allows gradient to pass through the originally un-differentiable sampling function.
The generative model described in section~\ref{sec:gen_model} then samples $z_{cat}$ and $z_{attr}$ from the distributions in order to generate an object that will be written to canvas $C_t$ using spatial attention module.

\subsection{Learning}
Similar to the original AIR model, we train Discrete-AIR model end-to-end by maximizing the lower bound on the marginal likelihood of data:

\begin{equation}\label{eq:objective}
log\, p_{\theta}(x) \leq \mathcal{L}(\theta,\phi) = \mathbb{E}_{q_{\phi}} \Big[\log \frac{p_{\theta}(x,z,n)}{q_{\phi}(z,n|x)}\Big]
\end{equation}

While in the original AIR model, one cannot further arrange this equation due to undifferentiable discrete variable sampling process used. For Discrete-AIR, by using Gubmel-Softmax as a repameterized sampling process, we can rearrange equation~\ref{eq:objective} as:

\begin{equation}\label{eq:objective_r}
\mathcal{L}(\theta,\phi) = \mathbb{E}_{q_{\phi}} \Big[\log p_{\theta}(x|z,n)\Big] - D_{KL}(q_{\phi}(z,n|x)||p(z,n))
\end{equation}
Where $p_{\theta}(x|z,n)$ is data likelihood and $D_{KL}$ is Kullback-Leibler (KL) divergence. This is the same implemented in the original VAE~\cite{kingma2013auto}.
Computing $\frac{\partial \mathcal{L}}{\partial \theta}$, the loss derivative with respect to parameters of the generative model, is relatively straightforward as it is fully differentiable. With a sampled batch of latent codes $z = (z_{cat},z_{attr},z_{where},z_{pres}) \sim q(\cdot|x) $, the partial derivative $\frac{\partial}{\partial \theta} p_{\theta}(x|z,n) $ can be directly computed. 

When computing $\frac{\partial \mathcal{L}}{\partial \phi}$, we can use the re-parameterization trick~\cite{kingma2013auto} to re-parametrise the sampling of both, continuous and discrete latent variables as a deterministic function in the form $h(\omega^i,\epsilon^i)$. $\omega^i$ is the parameters of the distributions for $z$ at time step $i$, and $\epsilon^i$ are random noise at time step $i$. In this way we can use the chain rule to compute the gradient with respect to $\phi$ as:

\begin{equation}
 \frac{\partial \mathcal{L}}{\partial \phi} =\frac{\partial \mathcal{L}}{\partial h} \times \frac{\partial h}{\partial \omega^i} \times \frac{\partial \omega^i}{\partial \phi}
\end{equation}

For our experiments, we parametrise continuous variables as multivariate Gaussian distributions with a diagonal covariance matrix. Thus $h(\omega^i,\epsilon^i) = \mu^i + \sigma^i * \epsilon^i$. For discrete variables, we use Gumbel softmax introduced in section~\ref{sec:discrete}, which is itself a re-parametrised differentiable sampling function.For the KL-divergence term, assuming all latent variables $z$ are conditionally independent, we can factorize $q(z|x)$ as $\prod_{i}q(z_i|x)$ and thereby separate the Kl-terms, as discussed in~\cite{dupont2018learning}. We use Gaussian prior for all continuous variables. While KL divergence between two Gumbel-Softmax distribution are not available in closed form, we approximate with a Monte-Carlo estimation of KL divergence with a categorical prior for $z_{cat}$, similar as~\cite{jang2016categorical}. For $z_{pres}$ we used a geometric prior and compute Monte-Carlo estimation of KL divergence~\cite{maddison2016concrete}.

\section{Evaluation}
We evaluate Discrete-AIR on two multi-object datasets, namely Multi-MNIST dataset as used in the original AIR model~\cite{eslami2016attend} and a multi-object shape dataset comprising of simple shapes similar to dSprites dataset~\cite{dsprites17}. We perform experiments to show that Discrete-AIR, while retaining the original strength of AIR model of discovering the number of objects in a scene, can additionally categorise each discovered object. In order to evaluate how accurately can Discrete-AIR categorize each object, we compute the correspondence rate between the best permutation of category assignments from Discrete-AIR model and the true labels of the dataset. 

To explain the metric we used, we first define a few notations. For each input image $x_i$, Discrete-AIR generates a corresponding category latent code $z^i_{cat}$ and presence variable $z^i_{pres}$. From this we can form a set of predicted object categories $O^i = \{o^i_1,\ldots,o^i_n\}$ for $n$ predicted objects where $o^i_k$ is the $k^{th}$ object category.
For each image we also have a set of true labels of existing objects $T^i = \{t^i_1,..t^i_m\}$. Due to non-identifiability problem of unsupervised learning where a simple permutation of best cluster assignment will give the same optimal result,
the category assignments produced by Discrete-AIR do not necessarily correspond to the labels. For example, an image patch of digit 1 could fall into category 4. We thus permute the category assignments and use the permutation that corresponds best with the true label as the category assignment. For example, for predicted category set $\{1,4,2,2,3\}$ and true label set $\{4,0,1,1,5\}$, we can use the following permutation of category for predicted category set $(1\rightarrow4,4\rightarrow0,2\rightarrow1,3\rightarrow5)$ to achieve best correspondence. To put it more formally, we define a function $f_p(C,p)$ where $C$ is a set or array of sets, and $p$ is an index permutation function to map elements in $C$. For the whole dataset, we have an array of predicted category set $O$ and an array of true label set $T$. We define correspondence rate as $in(O,T)/size(T)$ where $in(O,T)$ gives the number of true labels $t$ in $T$ that are correctly identified in $O$. $size(T)$ gives the total number of labels. We thus compute the best correspondence rate as:

\begin{equation}
R_{corr}=\max_{p\in P} \frac{in(O,T)}{size(T)} 
\end{equation}

where $P$ is the set of all possible permutations of predicted categories. This score is ranging from 0 to 1, and the score of a random category assignment should have expected score of $1/k$ where $k$ is the number of categories.

We train Discrete-AIR with the ELBO objective as presented in equation~\ref{eq:objective}. We use Adam optimiser~\cite{kingma2014adam} to optimise the model with batch size of 64 and learning rate of 0.0001. For Gumbel Softmax, we also applied temperature annealing~\cite{jang2016categorical} of $tau$ to start with a smoother distribution first and gradually approximate to discrete distribution. For more details about training, please see Appendix A in supplementary material. 

\subsection{Multi-Sprites}

To evaluate Discrete-AIR, we have built a multi-object dataset in similar style as the dSprites dataset~\cite{dsprites17}. This dataset consists of 90000 images of pixel size $64\times64$. In each image there are 0 to 3 objects with shapes in the categories of square, triangle and ellipse. The objects' spatial locations, orientations and size are all sampled randomly from uniform distributions. Details about constructing this dataset can be found in Appendix B. Figure~\ref{fig:multi-sprites} illustrates the application of Discrete-AIR on the Multi-Sprites dataset. Figure~\ref{fig:multi-sprites} a) shows samples of input data from the dataset with each object detected and categorised (with differently coloured bounding box). The number at the top-left corner shows the estimated number of objects in the scene. Figure~\ref{fig:multi-sprites} b) shows reconstructed images by the Discrete-AIR model. Figure~\ref{fig:multi-sprites} c)  shows the fully interpretable latent code of each object in the scene. For this dataset, we used a discrete variable of 3 categories as $z_{cat}$ together with spatial attention variables $z_{where}$. We did not include $z_{attr}$ for this dataset as the attributes of each object, including location, orientation and size, can all be controlled by  $z_{where}$. We did not include shear transformation $\mT^d_k$ in the spatial attention as the dataset generation process does not have a shear transformation. For more details about the architecture, please see Appendix A. 

\begin{figure}[h!]
	\centering
	\begin{subfigure}{\linewidth}
		\centering
		\includegraphics[height=1in]{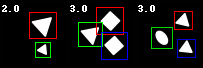}
		\caption{Data}
	\end{subfigure}%
	\\
	\begin{subfigure}{\linewidth}
		\centering
		\includegraphics[height=1in]{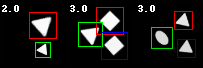}
		\caption{Reconstruction}
	\end{subfigure}
	\\
\begin{subfigure}{\linewidth}
	\centering
	\includegraphics[height=2in]{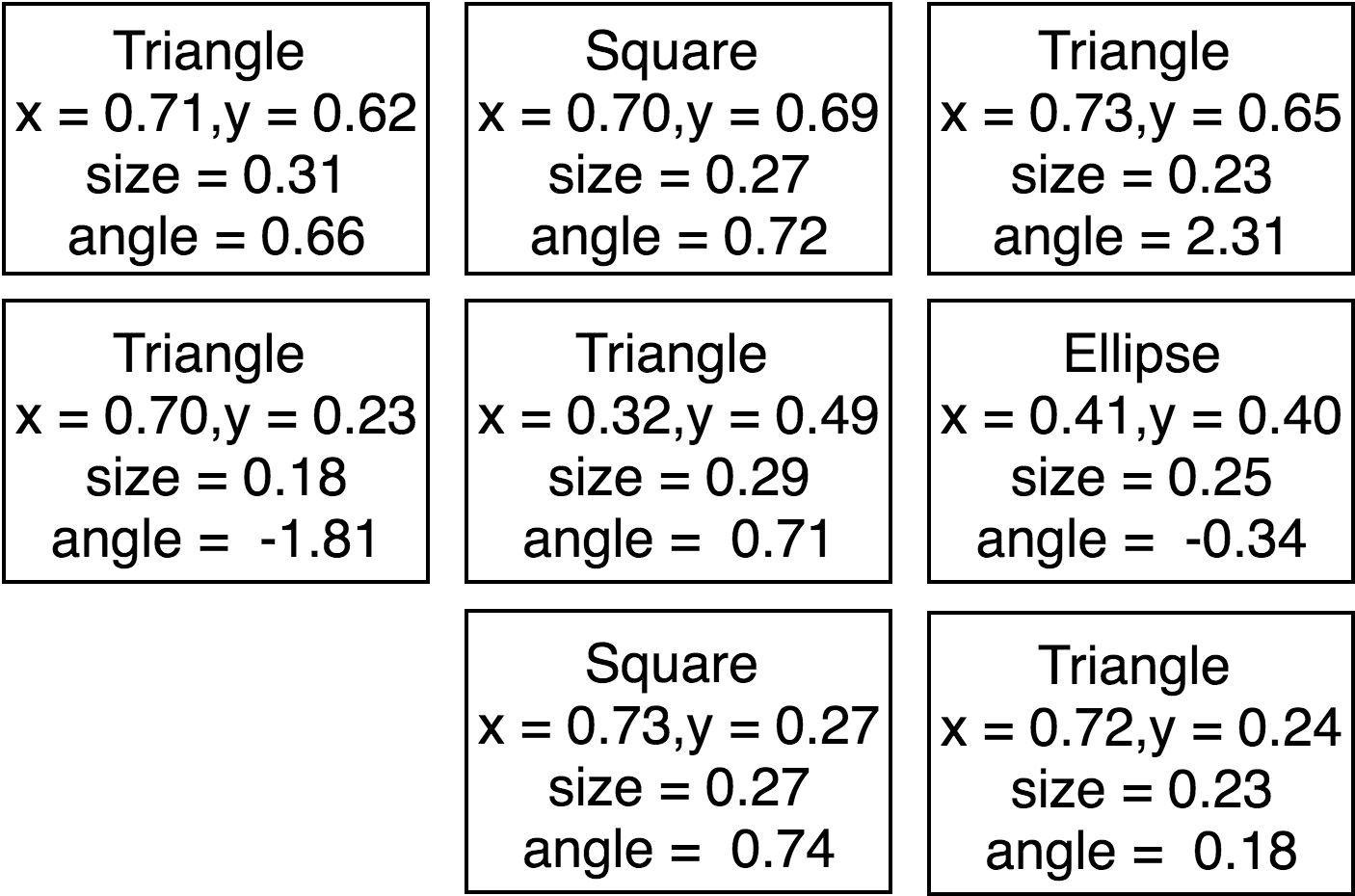}
	\caption{Latent codes}
\end{subfigure}
	\caption{Input data from Multi-Sprites dataset and reconstruction from Discrete-AIR model. The coloured bounding boxes show each detected object. The number at the top-left corner shows the count of number of objects in the image. Latent codes representing the scene, including object categories, sizes, spatial locations and orientation are also presented.}
	\label{fig:multi-sprites}
\end{figure}

For quantitative evaluation of Discrete-AIR we use three metrics, namely Reconstruction Error in the form of Mean Squared Error (MSE), count accuracy of number of objects in the scene and categorical correspondence rate. We also compare Discrete-AIR with AIR for the first two metrics. Table~\ref{tbl:multi_sprites} shows the performance for these three objectives. We report mean performance across 10 independent runs. Discrete-AIR has slightly better count accuracy than AIR, and is able to categorise objects with a mean category correspondence rate of 0.956. The best achieved correspondence rate is 0.967
Discrete-AIR does have increased reconstruction MSE compared to AIR model. However Discrete-AIR only uses a category latent variable of dimension 3, while the original AIR model uses 50 latent variables.
\begin{table}[h!]
  \begin{center}
	\begin{tabular}{||c | c c c ||} 
		\hline
		Model  & MSE & count acc. & category corr. \\ [0.5ex] 
		\hline\hline
		Discrete-AIR  & 0.096 & 0.985 & 0.945 \\ 
		\hline
		AIR  &0.074 & 0.981 & N/A \\ [1ex] 
		\hline
	\end{tabular}
    \caption{Quantitative evaluation of Discrete-AIR and comparison with AIR model for Multi-Sprites dataset.}
    \label{tbl:multi_sprites}
  \end{center}
\end{table}

\begin{figure}[h!]
	\begin{center}
		\includegraphics[width=\linewidth]{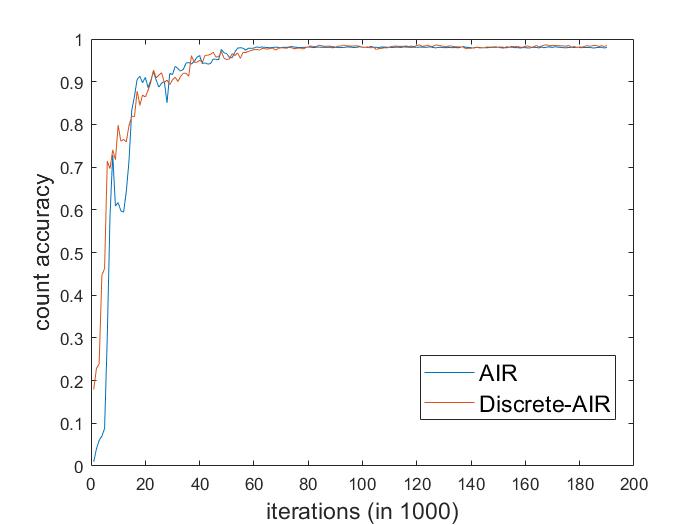}
	\end{center}
	\caption{Plot of count accuracy for AIR and Discrete-AIR during training for Multi-Sprites dataset.}
	\label{fig:sprites_count}
\end{figure}

We also plot count accuracy during training for both Discrete-AIR and AIR in figure~\ref{fig:sprites_count}. One can observe that both models converge towards similar accuracies, but Discrete-AIR model has slightly better increase rate and stability at the start of training.

\begin{figure}[h!]
	\centering
	\begin{subfigure}{0.24 \linewidth}
		\centering
		\includegraphics[height=2.4in]{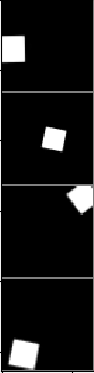}
		
	\end{subfigure}%
	\begin{subfigure}{0.24\linewidth}
		\centering
		\includegraphics[height=2.4in]{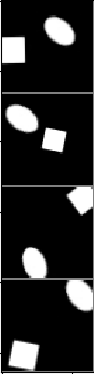}
		
	\end{subfigure}
	\begin{subfigure}{0.24\linewidth}
		\centering
		\includegraphics[height=2.4in]{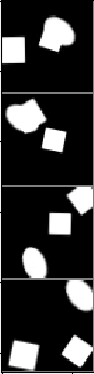}
	\end{subfigure}
	\begin{subfigure}{0.24\linewidth}
		\centering
		\includegraphics[height=2.4in]{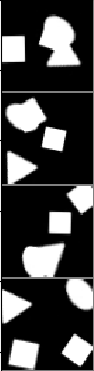}
	\end{subfigure}
	\caption{Generation of images by Discrete-AIR.}
	\label{fig:sprites_gen}
\end{figure}

Discrete-AIR can generate a scene with a given number of objects in a fully controlled way. We can specify categories of objects with $z_{cat}$ and their spatial attributes with $z_{attr}$. Figure~\ref{fig:sprites_gen} shows a sampled generation process. Note that while the training data contains up to 3 objects, Discrete-AIR can generate an arbitrary number of objects in the generative model. We generate 4 objects in the sequence "square, ellipse, square, triangle" with specified locations, orientation and size.

\subsection{Multi-MNIST}
\begin{figure}[h!]
	\centering
	\begin{subfigure}{\linewidth}
		\centering
		\includegraphics[height=1in]{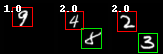}
		\caption{Data}
	\end{subfigure}%
	\\
	\begin{subfigure}{\linewidth}
		\centering
		\includegraphics[height=1in]{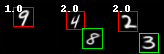}
		\caption{Reconstruction}
	\end{subfigure}

    \begin{subfigure}{\linewidth}
	    \centering
	    \includegraphics[height=2.2in]{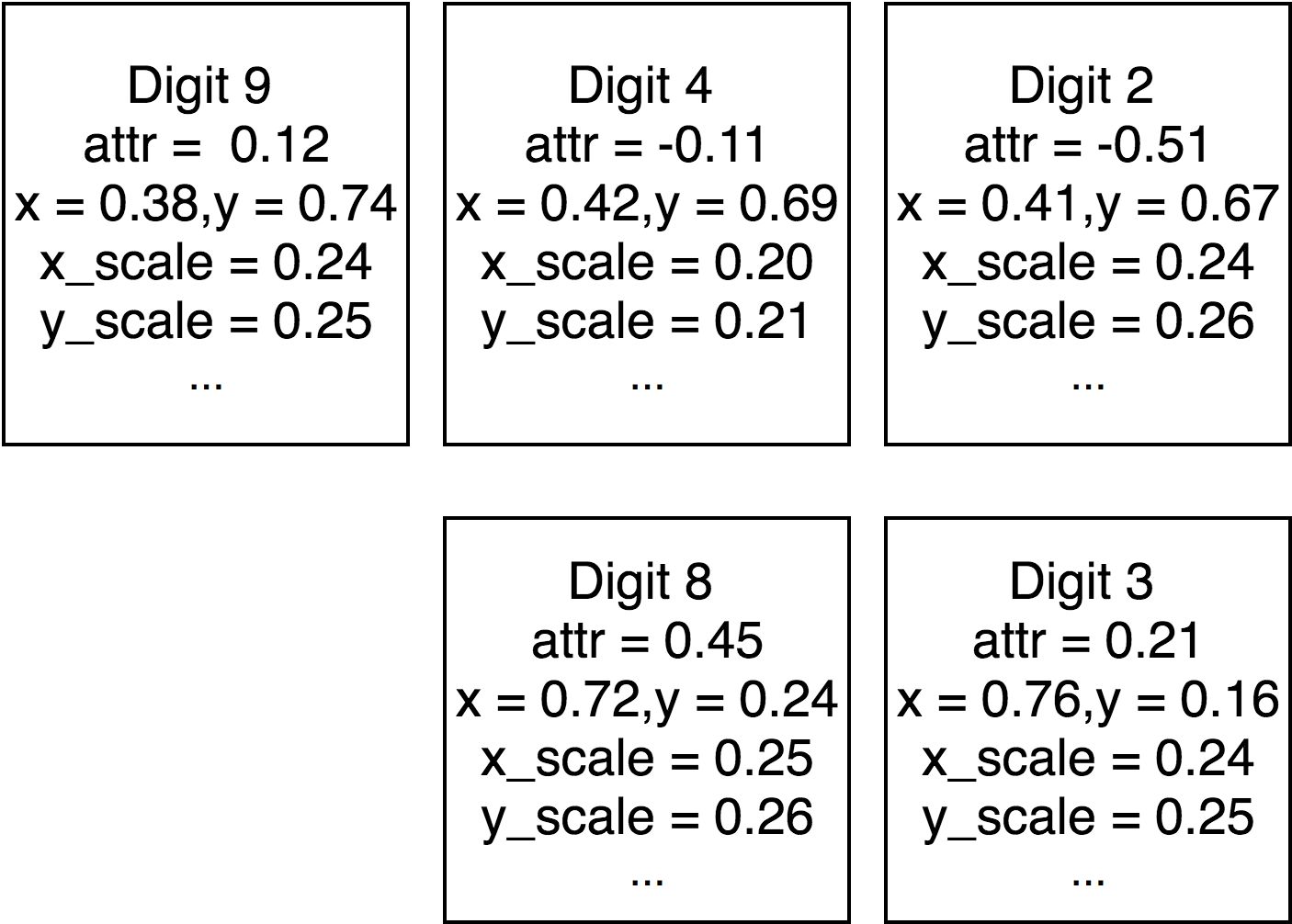}
	    \caption{Latent codes}
    \end{subfigure}
	\caption{Input data from Multi-MNIST dataset and reconstruction from Discrete-AIR model. The coloured bounding boxes show each detected object. The number at the top-left corner shows the count of number of objects in the image. Latent codes representing the scene, including digit categories, attribute variable value, sizes and spatial locations are also presented.}
	\label{fig:multi_mnist}
\end{figure}

We also evaluated Discrete-AIR on the Multi-MNIST dataset used by the original AIR model~\cite{eslami2016attend}. The dataset consists of 60000 images of size $50\times50$. Each image contains 0 to 2 digits sampled randomly from MNIST dataset~\cite{lecun1998gradient} and placed at random spatial positions. The dataset is publicly available in 'observations' python package\footnote{\url{https://github.com/edwardlib/observations}}. For this dataset, we choose a categorical variable with 10 categories as $z_{cat}$ and 1 continuous variable with Normal distribution as $z_{attr}$ as this gives best correspondence rate performance. We choose to combine transformation matrices $T_r$ and $T_k$ as one because this gives slightly better results. Figure~\ref{fig:multi_mnist} shows sampled input data from the dataset and reconstruction by Discrete-AIR. Figure~\ref{fig:multi_mnist} c) also show interpretable latent codes for each digit in the image. From this figure we can observe clearly that Discrete-AIR learns to match templates of category $z_{cat}$ with modifiable attributes  $z_{attr}$ to input data. For example, in the second image of input data, the digit '8' is written in a drastically different style from most other '8' in the dataset. However, as we can see in the reconstruction, Discrete-AIR is able to match a template of digit '8' with modified attributes such as slantedness and stroke thickness.

\begin{figure}[h]
	\begin{center}
		\includegraphics[width=\linewidth]{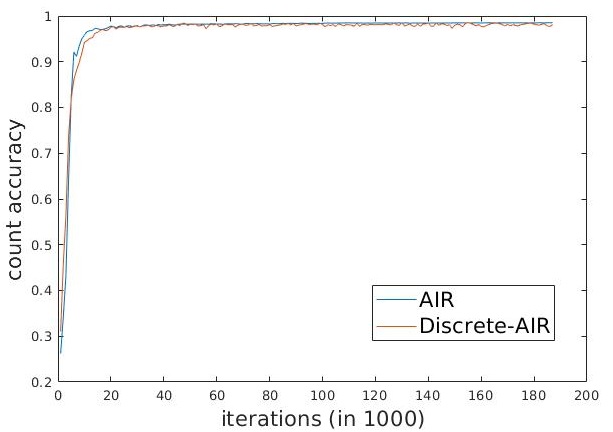}
	\end{center}
	\caption{Plot of count accuracy for AIR and Discrete-AIR during training for Multi-MNIST dataset.}
	\label{fig:mnist_count}
\end{figure}

We also performed the same quantitative analysis from Multi-Sprites dataset, as shown in table~\ref{tbl:multi_mnist}. For correspondence rate measurements, we only use 10\% subsampled data because permuting 10 digits requires $9!$ steps of evaluating across the dataset, which is too slow for the whole dataset.  While the count accuracy of Discrete-AIR and AIR model are very close, Discrete-AIR is able to categorize the digits in the image with a mean correspondence rate of 0.871. The best achieved correspondence rate is 0.913. We also plotted the count accuracy during training history, as shown in figure~\ref{fig:mnist_count}. We can see that Discrete-AIR's count accuracy increase rates are very close to that of the AIR model.

\begin{table}[h!]
	\begin{center}
		\begin{tabular}{||c | c c c||} 
			\hline
			Model & MSE & count acc. & category corr. \\ [0.5ex] 
			\hline\hline
			Discrete-AIR & 0.134 & 0.984 & 0.871 \\ 
			\hline
			AIR & 0.107 & 0.985 & N/A \\ [1ex] 
			\hline
		\end{tabular}
	\end{center}
	\caption{Quantitative evaluation of Discrete-AIR and comparison with AIR model for Multi-MNIST dataset.}
	\label{tbl:multi_mnist}
\end{table}

\begin{figure}[h!]
	\centering
	\begin{subfigure}{0.23 \linewidth}
		\centering
		\includegraphics[height=2.4in]{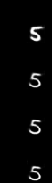}
		
	\end{subfigure}%
	\begin{subfigure}{0.23\linewidth}
		\centering
		\includegraphics[height=2.4in]{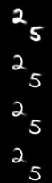}
		
	\end{subfigure}
    \quad
	\begin{subfigure}{0.23\linewidth}
		\centering
		\includegraphics[height=2.4in]{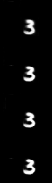}
	\end{subfigure}
	\begin{subfigure}{0.23\linewidth}
		\centering
		\includegraphics[height=2.4in]{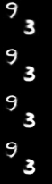}
	\end{subfigure}
	\caption{Generation of images by Discrete-AIR.}
	\label{fig:mnist_gen}
\end{figure}

Same as shown for Multi-Sprites dataset, Discrete-AIR is able to generate images in a fully controlled way with given categories of digits $z_{cat}$, attribute variable $z_{attr}$ and spatial variable $z_{where}$. Figure~\ref{fig:mnist_gen} shows a sampled generated image. Two digits are generated in subsequent images with attribute variable increasing from top to bottom. In the first sequence we generate digits '5' and '2' while in the second sequence we generate digits '3' and '9'. We can observe that the learnt attribute variable $z_{attr}$ encodes attributes that cannot be encoded by affine transformation spatial variable $z_{where}$. For example, increasing $z_{attr}$ increases the size of the hook space in digit '5', the hook space in digit '2', and the hook curve in digit '9'.

\section{Related work}

Generative models for unsupervised scene discovery have been an important topics in Machine Learning. Variational Auto-Encoder (VAE)~\cite{kingma2013auto} is one import generative model that learns in unsupervised way how to encode scenes into a set of latent codes that represent the scene at high-level feature abstraction. Among these unsupervised models, the DRAW model~\cite{gregor2015draw} combines Recurrent Neural Networks, VAE and attention mechanism to allow the generative model to focus on one part of image at a time, mimicking the foveation of human eye. AIR model~\cite{eslami2016attend} extends this idea to allow the model to focus on one integral part of the scene at a time, such as a digit or an object. 

One important topic in unsupervised learning is improving interpretability of learnt representations. For VAE, there have been various approaches in disentangling the latent code distribution, such as beta-VAE models~\cite{higgins2017beta,chen2018isolating,kim2018disentangling} which change the balance between reconstruction quality and latent capacity with $\beta$ parameters. The most related work in terms of disentangling is by Dupont et al~\cite{dupont2018learning}, which also disentangles latent code into discrete and continuous parts. However, this work can only disentangle for image that contains a single centered object, while our model works for scenes containing multiple objects.

Discrete-AIR extends AIR model to be able to not only discover integral parts, but assign interpretable, disentangled latent representations to these integral parts by encoding each parts into different categories and different attributes. Several works~\cite{eslami2016attend,wu2017neural,romaszko2017vision}, including AIR model itself, attempt to use pre-defined graphics like-decoder to generate sequences of disentangled latent representations for multiple parts of the scene. Discrete-AIR, to our best knowledge, is the first end-to-end trainable VAE model without a pre-defined generative function to  achieve this.

One other notable approach of disentangling scene representations is Neural Expectation Maximization (N-EM)~\cite{greff2017neural}, which develops an end-to-end clustering method to cluster pixels in an image. However N-EM does not have an encoder that encodes attended objects into latent variables such as object location, size and orientation, but only assigns pixels to clusters by maximising a likelihood function.

\section{Conclusion}
In summary, we developed Discrete-AIR, an unsupervised auto-encoder that learns to model a scene of multiple objects with interpretable latent codes. We have shown that Discrete-AIR can capture categories of each object in the scene and disentangle attribute variables from the categorical variable. Discrete-AIR can be applied on various problems where discrete representations are useful, such as on visual reasoning including solving Raving Progressive Matrices~\cite{barrett2018measuring} and symbolic visual question answering~\cite{yi2018neural}. These two works approach visual reasoning problems with supervised learning method where for each object its category, spatial parameters and attributes are labeled. Discrete-AIR can be used as a symbolic encoder or unsupervised pre-training of encoding model, thereby reduce or even completely remove the requirements for labeled data.

\bibliography{icml2019-duo-paper}
\bibliographystyle{icml2019}

\end{document}